# Tube-Balloon Logic for the Exploration of Fluidic Control Elements

Jovanna A. Tracz,† Lukas Wille,† Dylan Pathiraja, Savita V. Kendre, Ron Pfisterer, Ethan Turett, Gus T. Teran, Christoffer K. Abrahamsson, Samuel E. Root, Won-Kyu Lee, Daniel J. Preston, Haihui Joy Jiang, George M. Whitesides, and Markus P. Nemitz*

***Abstract*—The control of pneumatically driven soft robots typically requires electronics. Microcontrollers are connected to power electronics that switch valves and pumps on and off. As a recent alternative, fluidic control methods have been introduced, in which soft digital logic gates permit multiple actuation states to be achieved in soft systems. Such systems have demonstrated autonomous behaviors without the use of electronics. However, fluidic controllers have required complex fabrication processes. To democratize the exploration of fluidic controllers, we developed *tube-balloon logic* circuitry, which consists of logic gates made from straws and balloons. Each tube-balloon logic device takes a novice five minutes to fabricate and costs $0.45. Tube-balloon logic devices can also operate at pressures of up to 200 kPa and oscillate at frequencies of up to 15 Hz. We configure the tube-balloon logic device as NOT-, NAND-, and NOR-gates and assemble them into a three-ring oscillator to demonstrate a vibrating sieve that separates sugar from rice. Because tube-balloon logic devices are low-cost, easy to fabricate, and their operating principle is simple, they are well suited for exploring fundamental concepts of fluidic control schemes while encouraging design inquiry for pneumatically driven soft robots.**

## I. INTRODUCTION

Soft robotics is a relatively new field involving the use of soft materials with increased flexibility and adaptability when compared to those used in traditional, rigid robots. Soft actuators are particularly useful in more fragile applications (i.e., agriculture, healthcare) where soft actuators offer several advantages over their rigid counterparts including: (i) safety and compatibility with humans and animals [1], [2], (ii) low-cost [2], (iii) modularity [3], [4], (iv) the ability to conform to their surroundings [5], (v) high cycle lifetimes [6]–[8], and (vi) resistance to damage [9]. Because of such advantages, soft robots have gained traction in a variety of applications such as healthcare [10], [11], and the exploration of terrestrial environments [12], and are typically actuated using pneumatics [13], electrostatics [14], and magnetism [15], [16].

In the design of soft robots, soft materials are used for actuation, sensing, and structure, while rigid components are typically used for power, memory, and control. Rigid components used for the control of soft robotic actuators include pumps, microcontrollers, and valves, and require infrastructure such as power electronics and a computer for programming.

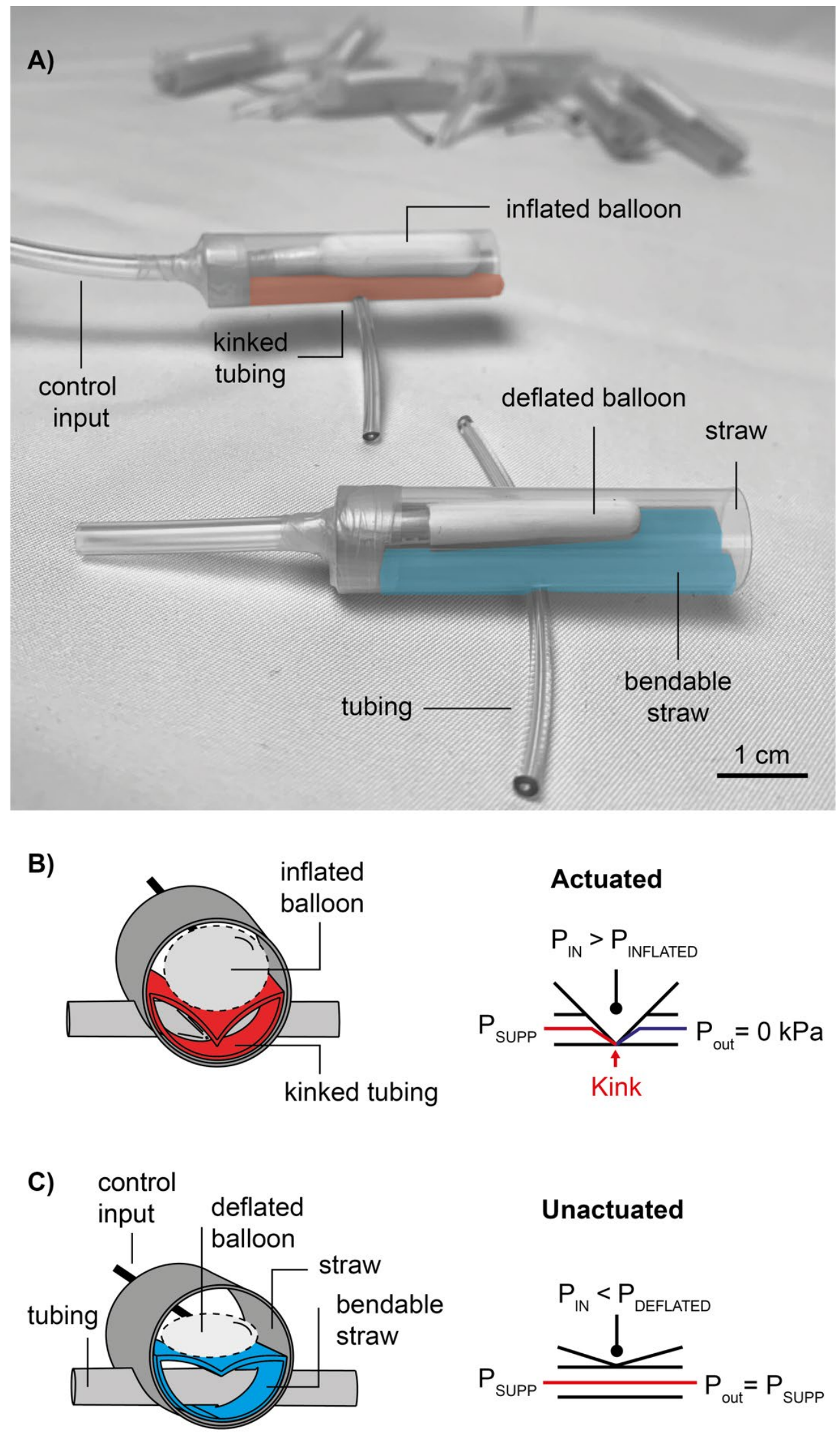

**Figure 1. Tube-balloon logic device.** The TBL device consists of straws, tubes, and a balloon. When the balloon is inflated, a bendable layer within the outer straw occludes air flow through polyvinyl chloride (PVC) tubing. A) Photo of TBL devices. The red TBL device is actuated, the blue TBL device is unactuated. B) Schematic in the actuated state: input pressure ($P_{IN}$) exceeds inflated pressure ($P_{INFLATED}$); the balloon inflates, kinks the tubing, and blocks pressure supply. C) Schematic in the unactuated state: input pressure ($P_{IN}$) is below deflated pressure ($P_{DEFLATED}$), balloon deflates, unblocking pressure supply.

This work was supported by the Department of Energy, Office of Basic Energy Science, Division of Materials Science and Engineering, grant ER45852, which funded work related to experimental apparatus and demonstrations. We would also like to acknowledge the Harvard NSF funded MRSEC (DMR-1425070) and funding from WPI for salary support of MPN.

Jovanna Tracz is with Eastern Virginia Medical School, VA, USA. Lukas Wille is with the Department of Chemistry at Oxford University, UK. Dylan Pathiraja, Christoffer Abrahamsson, Samuel E. Root, Won-Kyu Lee, Joy Haihui Jiang, and George M. Whitesides are with the Department of Chemistry and Chemical Biology at Harvard University, MA, USA. Daniel J. Preston is with the Department of Mechanical Engineering at Rice University, TX, USA. Savita V. Kendre, Ron Pfisterer, Ethan Turett, Gus T. Teran and Markus P. Nemitz are with the Department of Robotics Engineering at Worcester Polytechnic Institute, MA, USA.

† Indicates equal contribution
*To whom correspondence may be addressed: mnemitz@wpi.edu

There has been an increase in research effort related to the development of non-electronic control methods for soft robots, mostly in attempt to further improve the degree of integration and compliance of soft robotic systems. These soft controllers can be divided into microfluidic controllers, with channel sizes in the sub millimeter range, and macrofluidic controllers, with channel sizes in the millimeter range.

Wehner et al. demonstrated the first soft robotic system with integrated fluidic control [17]. A microfluidic oscillatory circuit was 3D printed and powered by a monopropellant. Octobot actuated a network of soft legs without the use of electronic components. Rothemund et al. developed a macrofluidic soft, bistable valve that can switch pressures in the range of 10 to 50 kPa [6]. A hemi-spherical membrane flips when a pneumatic signal is applied, switching air flow on and off. Preston et al. configured soft bistable valves as logic gates (AND, OR and NOT gates), enabling the development of soft digital logic for the first time [18]. Preston et al. also assembled three NOT gates into a ring oscillator, which generated an oscillatory output from a constant supply pressure, enabling undulating actuation schemes [8]. Nemitz et al. modified the membrane thickness of the soft bistable valve to develop non-volatile memory elements [19]. Drotman et al. used ring oscillators composed of soft valves to generate (1) walking gaits for a soft-legged quadruped using a single pressure supply source and (2) pneumatic memory for the soft robot to select between gaits [20].

Although micro- and macrofluidic control elements are non-electronic and require only one pressure supply to be operational (e.g., a pressure tank, generator, or $CO_2$ cartridge), their fabrication process is laborious; including the use of computers, 3D printers, vacuum chambers, and casting of elastomers. This may restrict the implementation of fluidic logic to universities or specialized makerspaces and laboratories.

In this paper, we introduce the tube-balloon logic (TBL) device, a soft logic gate that is made from tubing, balloons, and straws. The main contributions of this work include:

1. Development of a soft logic device that can perform binary logic functions (NOT-, NOR-, and NAND-gates) and be assembled into a ring oscillator that operates at high pressures (~200 kPa) and high oscillation frequencies (~15 Hz).
2. Assembly of a TBL device by a novice in ~5 minutes (instructions in **Movie S1**) and material costs of $0.45 (bill of materials in **Table 1**). The components of our device are inexpensive and widely available.
3. Demonstration of a vibrating sieve that is operated by a ring oscillator composed of three TBL devices. The sieve separates sugar from rice.

## II. Design

### A. Tube-balloon logic device

A TBL device consists of one straw cut into two shorter straws, a balloon, and flexible tubing. The outer straw acts as a housing chamber for the inner straw. The inner straw is a bendable layer that occludes air flow through the tubing upon compression by the inflating balloon (**Figure 1**).

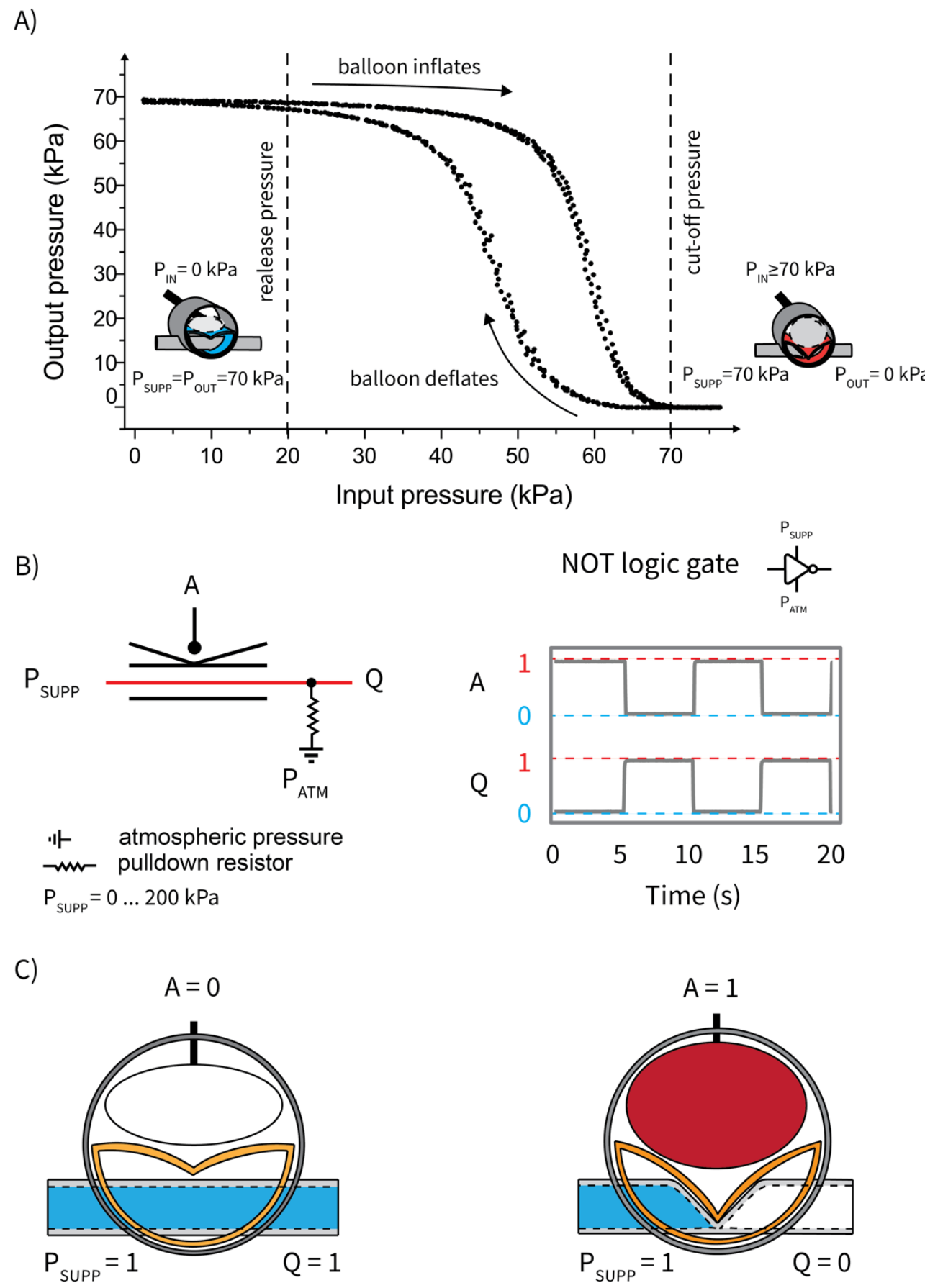

**Figure 2. TBL device configured as a NOT gate.** A) Graph depicting input and output pressures at which the balloon is inflated or deflated, kinking or unkinking the tubing to inhibit or permit pneumatic flow through the device. Supply pressure was steadily increased using a SMC ITV00 electronic regulator, and pressure measurements were recorded using Panasonic ADP5151 pressure sensors, both connected to a NI DAQ platform. B) Schematic depicts how output pressure varies when the device is actuated versus unactuated. The use of a pull-down resistor to allow for device deflation when unactuated is also displayed. C) Schematic representation of TBL device in unactuated (left) and actuated (right) states.

A TBL device functions as a single inverter. The tubing kinked by the bendable layer is connected to supply pressure, the balloon is connected to control pressure, and the pull-down resistor is connected to atmospheric pressure (**Figure 2**). When the balloon inside a TBL device inflates, it cuts off airflow through the tubing, and a pull-down resistor connected to atmospheric pressure releases the trapped air that has been cut off by the balloon. Each tube-balloon logic gate, configured as a NOT gate, requires ~85 kPa to switch states and can withstand gauge pressures up to ~200 kPa. We tested TBL devices for >200,000 actuation cycles at a frequency of 15 Hz. For detailed information on the fabrication process and materials used, please review the Methods section and our video (**Movie S1**).

A NOT gate is one of three basic Boolean operators; exemplifying the fundamental principles in circuit design that assigns all signals to either true (1) or false (0). TBL devices can also be stacked to implement NOR and NAND gates.

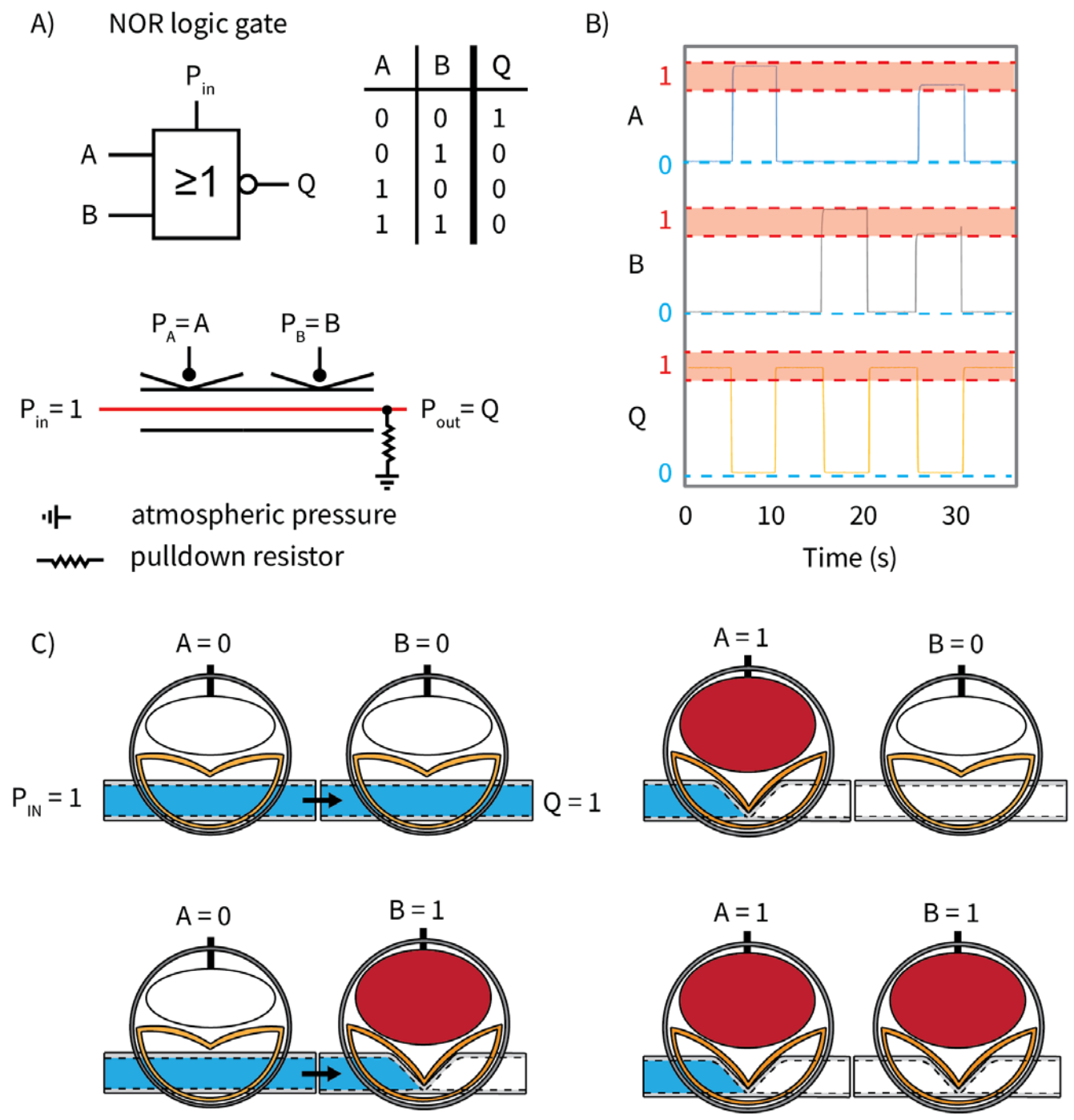

**Figure 3. TBL device configured as a NOR gate.** A) Truth table. B) Characterization of the implemented NOR gate. C) Schematic of the NOR-gate. When two TBL devices are connected in series, air is only able to pass (Q = 1) if both TBL devices have deflated balloons (A = B = 0). Otherwise, air is being cut off (Q = 0). The corresponding logic is equivalent to a NOR-gate. NOR-gates are functionally complete, that is, any combinational logic can be developed from combinations of NOR-gates.

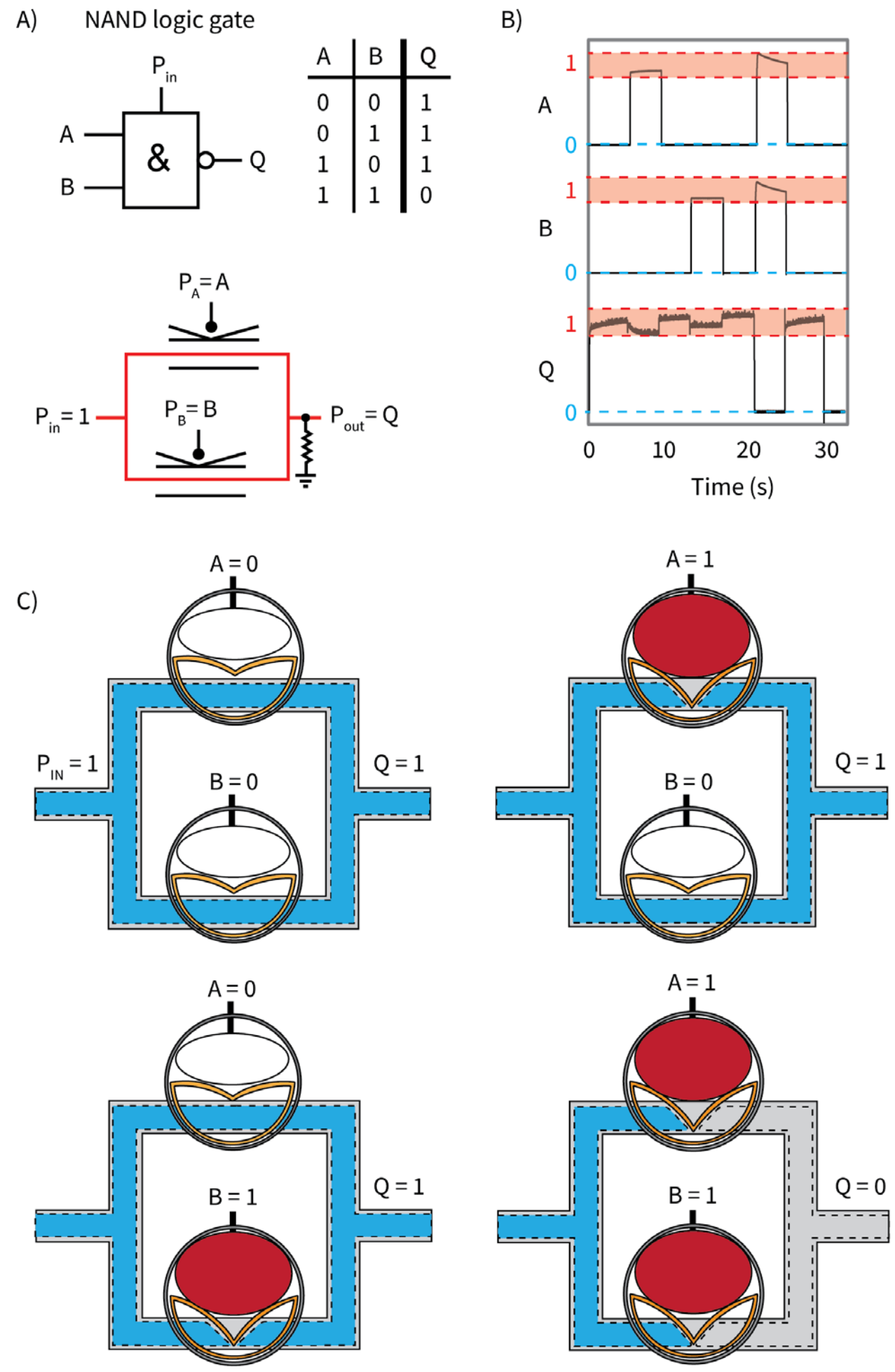

**Figure 4. TBL device configured as a NAND gate.** A) Truth table. B) Characterization of the implemented NAND gate. C) Schematic of the NAND-gate. When two TBL devices are connected in parallel, air can pass (Q = 1) when at least one TBL device has a deflated balloon (A = 0 or B = 0). Otherwise, air is being cut off (Q = 0). The corresponding logic is equivalent to a NAND-gate. NAND-gates are functionally complete, that is, any combinational logic can be developed from combinations of NAND-gates.

### *B. TBL devices configured as a NOR gate*

Two TBL devices can be connected in series to create a NOR-gate **(Figure 3).** A NOR gate is a combination of an OR-gate and a NOT-gate. It only outputs HIGH (Q = 1) when both pressure inputs are LOW (A = B = 0), which is the inversion of an OR gate. A NOR gate is visually easy to understand from the assembly of TBL devices; it suffices to inflate one balloon to block air passage through the NOR gate.

### *C. TBL devices configured as a NAND gate*

Two TBL devices can be connected in parallel to create a NAND-gate (**Figure 4**). A NAND gate is a combination of an AND-gate and a NOT-gate. It only outputs LOW (Q = 0) when both pressure inputs are HIGH (A = B = 1), which is the inversion of an AND gate. NOR- and NAND-gates are *functionally complete*, that is, any other logic function can be derived from them, which is an important concept in computer engineering.

### *D. TBL devices configured as a ring oscillator*

Odd numbers of NOT-gates can be connected in series to create a ring oscillator, whereas the last device must be connected to the first, forming a ring. A ring oscillator converts a constant pressure into oscillating pressures by inflating and deflating logic gates, which confers alternatingly unactuated and actuated states in a stable configuration. The oscillation frequency is dependent on a range of variables including the volume of the balloon that kinks off the tubing (a larger balloon takes longer to inflate than a smaller balloon), the material the balloon is composed of (an elastic material will force the air out of the balloon when disconnected from supply pressure), the length and diameter of the pull-down resistor (larger resistances deplete air slower), and the length of tubing connecting logic devices. We refer to our previous work for an in-depth analysis of factors contributing to ring oscillations [8].

In our case, three tube-balloon logic devices configured as NOT gates and assembled to a three-ring oscillator achieved an oscillation frequency of 15 Hz at a supply pressure of 145 kPa, and oscillation peak pressures of 35 kPa **(Figure 5**). The supply pressure is significantly larger than the peak pressure due to pressure dissipation through pull-down resistors. Supply pressure was measured prior to connecting the pneumatic load (i.e., ring oscillator).

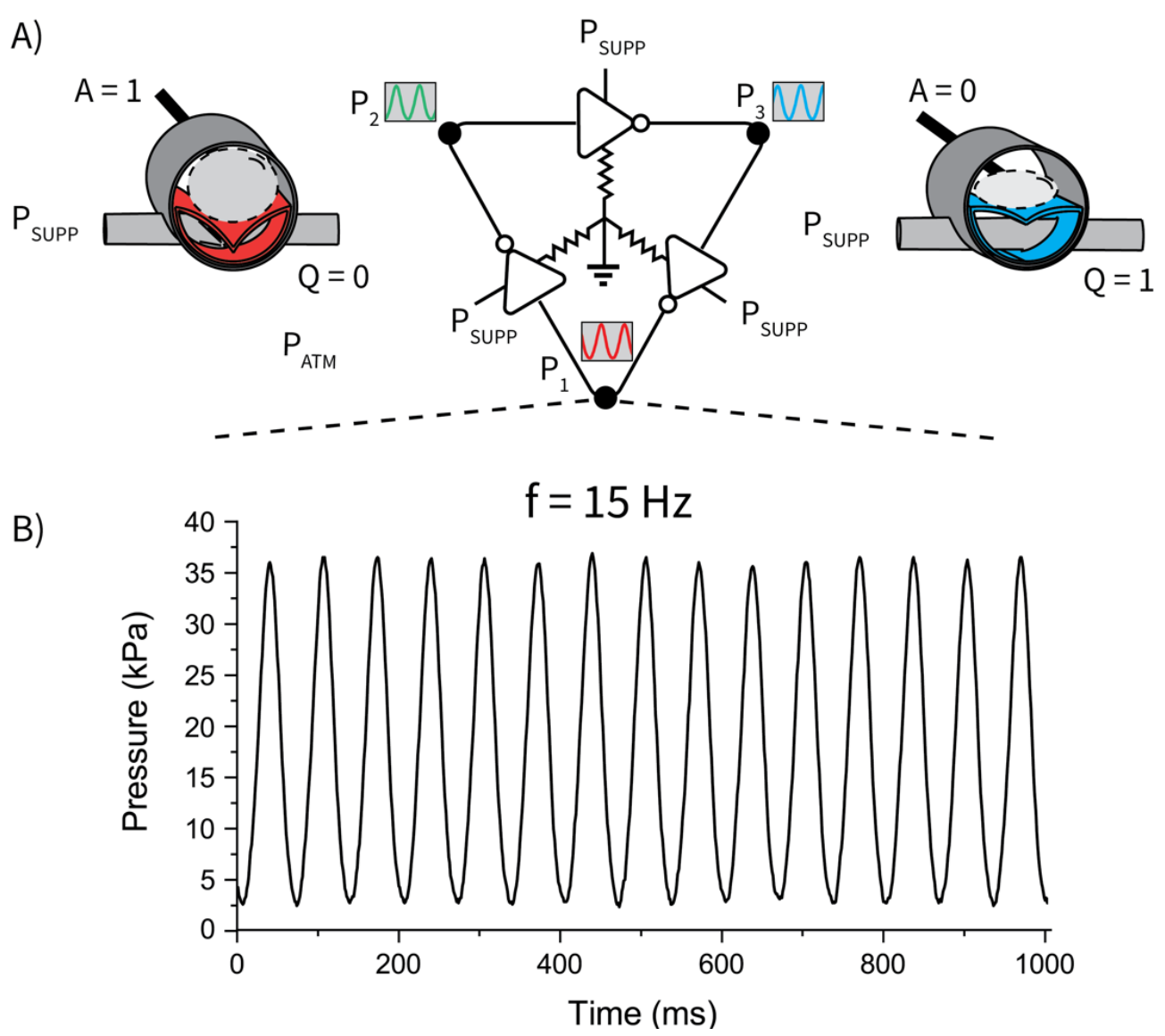

**Figure 5. TBL device as a ring oscillator.** A) Three TBL devices, configured as NOT gates, connected in series to function as a ring oscillator. A central pull-down resistor permits the balloon to fully deflate when unactuated. B) Pressure versus time graph reveals the TBL ring oscillates at a frequency of 15 Hz.

## III. Demonstration

Because of the relatively high oscillation frequency of 15 Hz achieved by the TBL ring oscillator, which is an order of magnitude higher than currently existing soft ring oscillators [8], we demonstrate the TBL ring oscillator as a pneumatic, vibrating sieve that is used for the separation of granular materials of varying grades (rice and sugar). To that end, we created a multi-staged device (**Figure 6)** that consists of a porous sieve (**Figure 6A**), a particle collection plate (**Figure 6B**), an actuation plate (**Figure 6C**), and a chamber containing the three-ring TBL oscillator (**Figure 6D**). The chamber has several inlets and outlets for incoming supply pressure and outgoing pull-down resistors and balloon actuators. The pull-down resistors exit the chamber of the ring oscillator (**Figure 6D**) and connect to the outer walls of the porous sieve (**Figure 6A**) to support particle separation through airflow. Three balloon actuators are connected to the three outputs of the ring oscillator (**Figure 6A**) and rest on the actuation plate (**Figure 6C**), inducing a vortex motion within the media during separation **(Figure 6E, Movie S1).** The porous filtering chamber permits the sugar, but not the rice to flow through, separating particles based on size. The separation process took 15 seconds and is detailed in the supplemental information (**Movie S1**).

## IV. Methods

### *A. Fabrication of a TBL device*

The fabrication process of a TBL device consists of cutting a straw with a diameter of 12.7 mm into two 40 mm pieces; folding one straw in the middle to create a bendable layer (with a heart-like shape); inserting the bendable layer into the outer straw; punching holes into both straws as an inlet for the PVC tubing using a hole puncher; and feeding a flexible PVC tube of 7.5 cm in length and with an inner diameter of 1 mm through both straws (bendable layer and outer straw); and inserting the balloon inside the outer straw such that it lays in between the outer straw and the bendable straw layer (**Figure 1**). Before inserting the balloon, we attached a 7.5 cm long PVC tube to it. We cut the balloon to 2 cm length, pushed it over the tubing with an overlap of 1 cm, and fixed it with Parafilm. We also applied parafilm to both ends of the TBL device, to secure the balloon in place, and to ensure that the balloon cannot overinflate beyond the straw and burst. We attached a pull-down resistor with an inner diameter of 1 mm and a length of 15 cm to the device. **Table 1** shows the bill of materials for a single TBL device. All materials are commonly available and can be readily acquired from Amazon and McMaster-Carr.

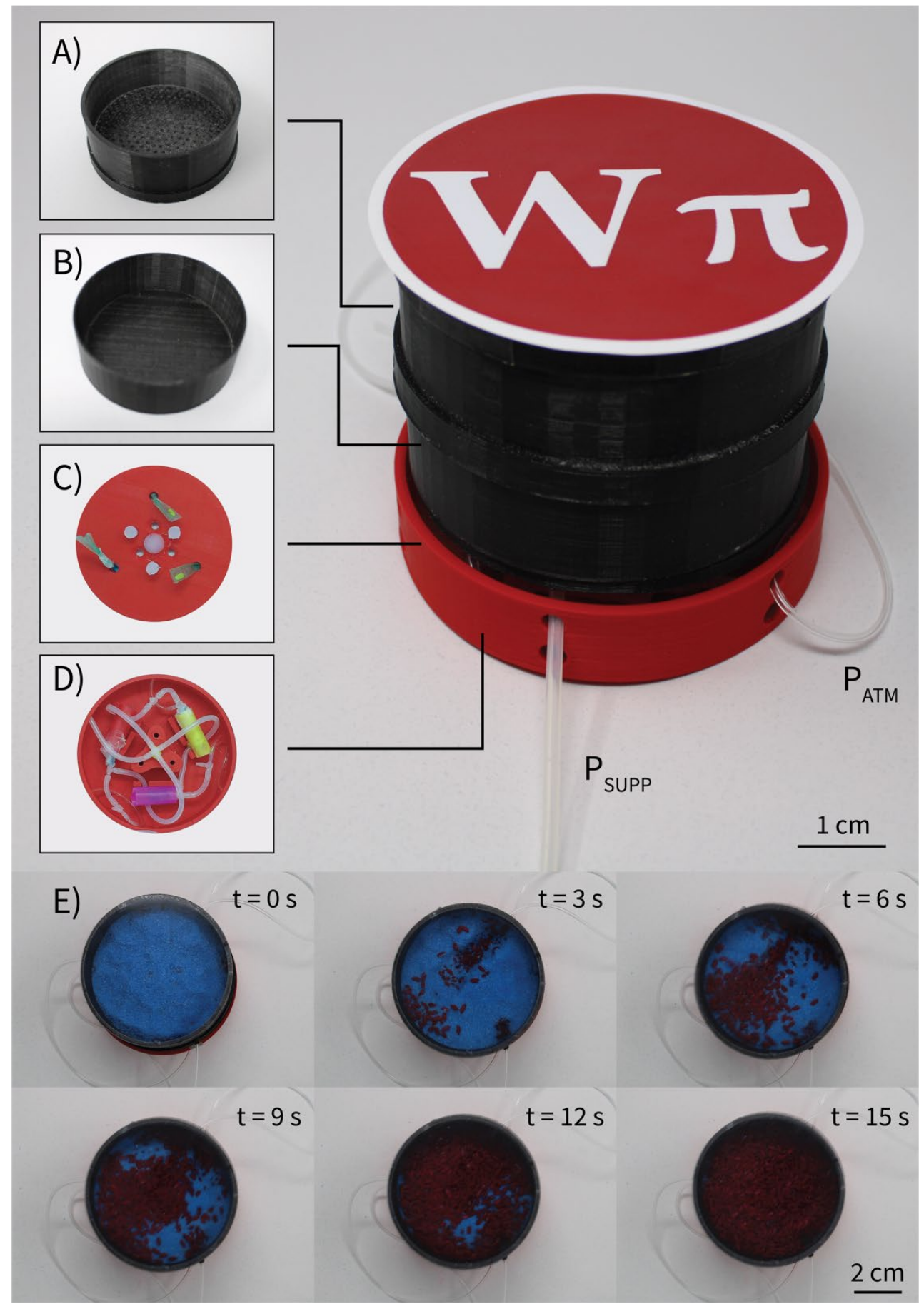

**Figure 6. TBL ring oscillator as a pneumatic vibrating sieve for the separation of granular from bulk media.** Three TBL devices, configured as NOT gates connected in a ring function as an oscillator. The pneumatic oscillator causes a plate to vibrate. On top of this vibration plate, rests a porous, filtering chamber which permits the sugar, but not the rice to flow through, separating particles based on size. A) Porous sieve for filtration. B) Particle collection plate. C) Actuation plate, including three balloons that inflate successively along the edges of the sieve, inducing a vortex motion within the media during separation (Movie S1). D) Chamber containing the ring oscillator. E) Time lapse images depicting sieving process (**Movie S1**). Blue: sugar, Red: rice. WPI logo (Worcester Polytechnic Institute).

| Description | Supplier | Unit | Cost |
|---|---|---|---|
| ALINK 100 1/2" (Boba straw) | Amazon.com | 1 | $0.08 |
| Koogel 260Q (Twisting balloon) | Amazon.com | 1 | $0.05 |
| PVC tubing (ID: 1 mm) | McMaster-Carr | 30 cm | $0.29 |
| Parafilm M | Amazon.com | 6 cm$^2$ | $0.03 |
| **Cost per TBL device** | | | **$0.45** |

**Table 1. Bill of Materials.** A single TBL device costs $0.45. Materials can be readily acquired from Amazon and McMaster-Carr. **Movie S1** explains the fabrication process in detail.

### *B. Fabrication of vibrating sieve*

We custom designed the vibrating sieve using Autodesk Inventor. We printed the porous sieve, particle collection plate, actuation plate, and the chamber containing the ring oscillator, using a FDM printer (Prusa MK3S) and Polyactic acid (PLA) as a filament.

### *C. Data acquisition and testing*

For the characterization of our TBL device, we used three pressure sensors (Panasonic ADP5151) and an electronic pressure regulator (SMC ITV001) connected to a data acquisition tool (NI USB-6009). A custom written LabView program was used to record the data. For the demonstration of separating sugar from rice, we colored both media with pigments from SmoothOn Inc. We used the pressure supply line in the laboratory at Worcester Polytechnic Institute to operate TBL devices. An electric pump also achieves sufficient pressures (12 V diaphragm pump, Amazon.com).

## V. Discussion

### *A. Pull-down resistors*

TBL devices provide an inexpensive method for exploring simple, fluidic Boolean logic functions. However, due to the use of pull-down resistors, we require a constant supply of pressure (e.g., from an electric pump), making mobile pressure sources such as $CO_2$ cartridges inapplicable.

TBL devices can be stacked but are ultimately limited by the airflow that escapes through the pull-down resistors. Pull-down resistors cause a pressure drop in supply pressure, and eventually the supply pressure will be insufficient to switch TBL devices ($P_{Supply}<P_{Control}$). In comparison, soft digital logic gates using soft bistable valves do not require pull-down resistors due to kinking only one out of two tubes at any given time (the other tube remains unkinked) [18]. Nevertheless, soft bistable valves are also more difficult to fabricate and operate at lower pressures (< 100 kPa).

### *B. Programmability*

Fluidic controllers lack the programmability of electronic controllers. While electronics can be easily re-programmed, fluidic controllers need to be manually re-arranged to implement different functionalities. Electronic controllers can also implement significantly more complex control algorithms than fluidic controllers (e.g., PID controllers). Fluidic controllers therefore possess the advantage of operating in the same physical domain as pneumatic soft robots: pressurized air. The electro-fluidic transduction using microcontrollers and valves is not required, potentially allowing for highly integrated fluidic designs.

### *C. Electronic control boards*

Exploring electronic control schemes for soft robots typically requires fluidic control boards which can cost hundreds of dollars (popular toolkits include: ProgrammableAir - $175; FlowIO - $520; Soft Robotics Toolkit - $800). In comparison, fluidic controllers from TBL devices, while less complex, are less costly to implement, with the opportunity to develop entirely fluidically operated systems. This may have promising applications in harsh environments including radioactive and explosive environments. Modern electronics are easily destroyed by ionizing radiation, and electronics for explosive environments undergoes a specialized certification process which can be costly.

### *D. Education*

Our technology may be interesting to educators as well. Although this work does not include a case study with a group of middle and high school aged students, we provide in depth details on the material choice and fabrication process (**Table 1** and **Movie S1**). We envision entry-level literature on digital logic to pair well with experiments using tube balloon logic. For example, AND- and OR-gates can be assembled from NOR and NAND gates. We refer to our previous work on soft bistable valves for other circuits including set-reset latches and shift registers for educators interested in applying this technology in the classroom [18].

## VI. Conclusion

We present a pneumatic logic device that operates at high pressures (~200 kPa), and high oscillation frequencies (~15 Hz). It can be fabricated by a novice in 5 minutes from low-cost materials (straws, balloons, tubing, and Parafilm), with a total cost of $0.45 per device. We configured tube balloon logic devices as NOT-, NOR-, and NAND-gates and assembled them to a three-ring oscillator. We demonstrate that our three-ring oscillator can operate a vibrating sieve for the separation of granular media.

The TBL device provides an inexpensive, simple method of exploring simple Boolean logic operations and potentially interconnecting them with existing soft robotic actuators. TBL devices showcase a method for students and researchers to apply fluidic control elements, as well as comprehend a fundamental understanding of Boolean logic without accessing specialized laboratories.